# Cognitive Robotics: For Never was a Story of More Woe than This

Emanuel Diamant
VIDIA-mant
Kiriat Ono , Israel
email : emanl.245@gmail.com

*Abstract*— **We are now on the verge of the next technical revolution – robots are going to invade our lives. However, to interact with humans or to be incorporated into a human "collective" robots have to be provided with some human-like cognitive abilities. What does it mean? – Nobody knows. But, robotics research communities are trying hard to find out a way to cope with this problem. Meanwhile, despite abundant funding these efforts did not lead to any meaningful result (only in Europe, only in the past ten years, Cognitive Robotics research funding has reached a ceiling of 1.39 billion euros). In the next ten years, a similar budget is going to be spent to tackle the Cognitive Robotics problems in the frame of the Human Brain Project. There is no reason to expect that this time the result will be different. We would like to try to explain why we are so unhappy about this.**

## I.  INTRODUCTION

From the beginning, it was a fascinating idea: to create human-like living beings that would help and assist us in our tedious everyday duties. The history has preserved many famous stories about such undertakings – Pygmalion and his Galatea, Talos the guard of Crete (both from the ancient Greek mythology), Maharal's Golem from the late 16th century Prague, Frankenstein's monster of the early 19th century.

In the year 1920, that was Karel Capek who gave them their present-day name – Robots. In 1942, Isaac Azimov was the first who introduced the term – Robotics. In 1959, the first real, not imagined and not legendary, industrial robot had entered the factory floor and, strictly speaking, has heralded the beginning of the robotics era. Then, at the end of the past century, robots start to appear in our human surroundings.

It has immediately become clear that, to work with humans (in cooperation and in tight interaction with them), robots have to possess some human-like cognitive abilities. What does it mean "to possess human-like cognitive abilities"? – Nobody knew then, nobody knows today. But that does not matter – the robotics research community enthusiastically started to cope with the challenge, endorsed with ample budget funding provided by the USA Defence Advanced Research Projects Agency (DARPA) and the European Union Research and Technological Development (EU RTD) programme. We would like to provide a short account of these efforts.

## II.  MARCH TO THE GLORIOUS FUTURE

As it was just said above, the DARPA in USA and the European Commission in Europe are today the most prominent supporters of scientific and technological progress, which are operating worldwide and are promoting an extensive range of critically important research initiatives. In the past 10-15 years, Cognitive Robotics was certainly one among them.

### A.  DARPA's Projects on Cognitive Robotics

The DARPA has always posited itself as an authority aimed to address a wide range of technological opportunities directed to meet the national security challenges. Endorsed with a budget of up to $2.8 billion (in FY 2013), it pursues its objectives through a wide range of R&D programs [1]. Cognitive Robotics does not appear in DARPA's programme as a bundle of programs grouped by a common theme; on the contrary, in DARPA's practice Cognitive Robotics is handled as a collection of separate programs that share a common target issue. The list of Cognitive Robotics and Cognitive-Robotics-related programs launched in the years 2001-2013 can be seen in Table I.

DARPA's efforts on robotics are focused primarily on military and defence-related applications with a clear goal to bring real-time, integrated, multi-source intelligence to the battlefield. DARPA does not strive to replace the warrior with a robot, but it believes that it is possible to improve the abilities of individual warfighter by combining technological achievements with human brain cognitive capacities thus making information understanding and decision-making far more effective and efficient for military users. So, it tries hard, on one hand, to revolutionize the underlying technologies (for unmanned sensor systems and battlefield information gathering) and, on the other hand, to merge them with the next generation computational systems that will have some human-like cognitive capabilities (such as reasoning and learning capabilities) and levels of autonomy which are beyond those of the today's systems. The spectrum of programs presented in Table I reliably represents this DARPA's approach to Cognitive Robotics R&D.

TABLE I. DARPA'S PROJECTS ON COGNITIVE ROBOTICS

| | |
|---|---|
| Cognitive Computing Systems (CoGS) | 2008 – …….. |
| DARPA's Neovision project (NEOVISION2) | 2009 – …… |
| Video and Image Retrieval and Analysis Tool (VIRAT) | 2011 – …… |
| Autonomous Robotics Manipulation Program (ARM) | 2011 – …… |
| DARPA Robotics Challenge (DRC) | 2012 – …… |
| DARPA's Insight Program (DIP) | 2013 – …….. |
| Biologically Inspired Cognitive Architectures (BICA) | 2005 - 2007 |
| The Cognitive Technology Threat Warning System (CT2WS) | 2007 – …… |
| Cognitive Assistant that Learns and Organizes (CALO) | 2003 - 2008 |
| Personalized Assistant that Learns (PAL) | 2003 - 2008 |
| Augmented Cognition (AugCog) | 2001 - 2006 |

*B. The European programs on Cognitive Robotics*

European Union research is conducted in a frame of research programmes called Framework Programmes for Research and Technological Development, in short Framework Programmes farther abbreviated as FP1 to FP8.

Cognitive Robotics related issues start to appear in the FP5 programme and then, respectively, continue to evolve and expand in the following FP6 and FP7 work programmes. Contrary to DARPA's approach, EU R&D activities are clustered to several main "themes" that are further segmented into "challenges", which are further divided into "objectives" in frame of which the individual projects are carried out. Cognitive Robotics in the Frame Programmes FP6 and FP7 is represented as a Challenge (Challenge 2) of the Information and Communication Technologies (ICT) theme (Theme 3). At the time of the transition from FP5 to FP6, when subdivision to Challenges was not yet introduced, Cognitive Robotics and its related issues such as Cognitive Vision and four other particular items appear straight as objectives in the Information Society Technologies (IST) theme (see Table II).

TABLE II. ROBOTICS IN FP5

| Area | No. of projects | Total cost (M€) | Total EC funding (M€) |
|---|---|---|---|
| IST Demining | 8 | 30,6 | 15,6 |
| IST FET Neuro-IT | 15 | 32,4 | 23,1 |
| IST FET General | 17 | 39,2 | 25,7 |
| IST Cognitive Vision | 8 | 24,2 | 17,3 |
| GROWTH etc. | 24 | 63,2 | 34,9 |
| Total | 72 | **189,7** | 116,5 |

TABLE III. COGNITIVE ROBOTICS IN FP6

| Year | Objective | Total cost (M€) |
|---|---|---|
| 2002 Work Programme | IST2002 - IV.2.1 Cognitive vision systems<br>IST2002 - VI.2.2 Presence Research: Cognitive sciences and future media | ?<br>? |
| 2003-2004 Work Programme | 2.3.1.7 Semantic-based Knowledge Systems<br>2.3.1.8 Networked Audiovisual systems and home platforms<br>2.3.2.4 Cognitive Systems<br>2.3.4.2.(vii) : Bio-inspired Intelligent Information Systems<br>Proactive initiatives (i) Beyond robotics | 55<br>60<br>25 |
| 2005-2006 Work Programme | 2.4.6 Networked Audio Visual Systems and Home Platforms<br>2.4.7 Semantic-based Knowledge and Content Systems<br>2.4.8 Cognitive Systems<br>2.4.11 Integrated biomedical information for better health | 63<br>112<br>45<br>75 |
| | **Total** | **435** |

TABLE IV. COGNITIVE ROBOTICS IN FP7

| Year | Call | Objective | Budget (M€) |
|------|------|-----------|-------------|
| 2007 | Call 1 | ICT-2007.2.1 Cognitive systems, interaction, robotic<br>ICT-2007.4.2 Intelligent content and semantics<br>ICT-2007.8.3 Bio-ICT convergence | 96<br>51<br>20 |
| 2008 | Call 3 | ICT-2007.2.2 Cognitive systems, interaction, robotics<br>ICT-2007.4.3 Digital libraries and technology-enhanced learning<br>ICT-2007.4.4 Intelligent content and semantics<br>ICT-2007.8.5 Embodied Intelligence | 97<br>50<br>?<br>? |
| 2009 | Call 4 | ICT-2009.2.1: Cognitive Systems and Robotics<br>ICT-2009.2.2: Language-Based Interaction | 73<br>26 |
| 2010 | Call 6 | ICT-2009.2.1: Cognitive Systems and Robotics<br>ICT-2009.2.2: Language-Based Interaction | 80<br>30 |
| 2011 | Call 7 | ICT-2011-7 Cognitive Systems and Robotics | 73 |
| 2012 | Call 9 | ICT-2011-9 Cognitive Systems and Robotics | 82 |
| 2013 | Call 10 | ICT-2013.2.1 Robotics, Cognitive Systems & Smart Spaces, Symbiotic Interaction<br>ICT-2013.2.2 Robotics use cases & Accompanying measures | 67<br>23 |
| | | **Total** | **768** |

Juxtaposing Table II, Table III, and Table IV, it can be seen how from a Cognitive Vision objective in FP5 (a Cognitive Robotics related topic) Cognitive Robotics has evolved to a full-blown Challenge (Challenge 2) in the FP6 and FP7 programmes, steadily growing from 190 M€ in FP5 to near 1.2 billion euros in the next FP6 and FP7 programmes (435 M€ for FP6 + 768 M€ for FP7).

## III. DECEIVED EXPECTATIONS

During all these times, the declared goals of Cognitive Robotics programmes were: (As it follows from DARPA's News Releases) "to create adaptable, integrated intelligence systems aimed to augment intelligence analysts' capabilities to support time-sensitive operations on the battlefield" [11]. And in another document – "The objectives (of DARPA's programs) are not to replace human analysts, but to make them more effective and efficient by reducing their cognitive load and enabling them to search for activities and threats quickly and easily" [12].

Objectives of Challenge 2 programs in the EU research initiative have been far more ambitious – The FP6 Workprogramme for years 2003-2004 states: (The objective is) "to construct physically instantiated or embodied systems that can perceive, understand (the semantics of information conveyed through their perceptual input) and interact with their environment, and evolve in order to achieve human-like performance in activities requiring context-(situation and task) specific knowledge, etc. The development of cognitive robots whose "purpose in life" would be to serve humans as assistants or "companions". Such robots would be able to learn new skills and tasks in an active open-ended way and to grow in constant interaction and co-operation with humans" [4].

These objectives (almost in similar words) are repeatedly declared in all further Work programmes. For example, the 2011-2012 Workprogramme says that in these words: "Challenge 2 focuses on artificial cognitive systems and robots that operate in dynamic, nondeterministic, real-life environments… Actions under this Challenge support research on engineering robotic systems and on endowing artificial systems with cognitive capabilities" [8].

Careful examination of the outcome that results from both the DARPA's programs and from the FP5-FP7 objectives leads to a univocal conclusion – the announced goals of all these programs have never been reached!

The explanation of this phenomenon is very simple – people try to provide robots with human-like cognitive abilities, but at the same time the same people are devoid of even a slightest understanding about what does the notion of "human-like cognitive abilities" really mean.

During the past years, the problem has become obvious and has been even mentioned in the 2011-2012 Workprogramme: "Hard scientific and technological research issues still need to be tackled in order to make robots fit for rendering high-quality services, or for flexible manufacturing scenarios. Sound theories are requisite to underpinning the development of robotic systems and providing pertinent design paradigms, also informed by studies of natural cognitive systems (as in the neuro- and behavioural sciences) [8].

Even more definite was the statement of the year 2013 Work programme – "An additional research focus targeted under this challenge will address symbiotic human-machine relations, which aims at a deeper understanding of human behaviour during interaction with ICT, going beyond conventional approaches. The work on cognitive systems and smart spaces and on symbiotic human-machine relations is not restricted to robotics" [13].

This promise was also left unfulfilled. At the end of 2013, Cognitive Robotics research has moved to and has tightened itself with the human brain research activities.

## IV. NEW HOPES

At the beginning of year 2014, both Europe and USA will launch ambitious programmes for human brain research. In the USA, the programme is called the Brain Research through Advancing Innovative Neurotechnologies (BRAIN) Initiative and it was announced by President Barack Obama on April 2013. Its accomplishment will be led by the National Institutes of Health (NIH), DARPA, and the National Science Foundation (NSF) [14].

In Europe, the Human Brain Project is a ten-year project, consisting of a thirty-month ramp-up phase, funded under FP7, with support from a special flagship ERANET, and a ninety-month operational phase, to be funded under Horizon 2020 programme. The project, which will have a total budget of over 1 billion Euros, is European-led with a strong element of international cooperation. The goal of the project is to build a completely new ICT infrastructure for neuroscience, and for brain-related research in medicine and computing, catalysing a global collaborative effort to understand the human brain and its diseases and ultimately to emulate its computational capabilities [15].

The main features of the two projects are collected in the Table V.

As it follows from the Table V, Cognitive Robotics is not among the main goals of the two Flagship initiatives, but it is definitely among their main purposes. In the European Human Brain Project it appears as the "Cognitive Architectures" line in the list of the HBP topics. In the American BRAIN Project Cognitive Robotics issues are hidden behind the "Link neuronal activity to behaviour" topic.

## V. AN ATTEMPT TO PREDICT THE FUTURE

In attempt to predict the future results of these two projects, let us juxtapose them with something that is well known to us and that we are quite familiar with. We mean the enduring and persistent study of National Economics. While the human nervous system can be seen as the driving force behind the behaviour of a single human, national economics can be seen as the driving force behind the behaviour of a whole human society. Both are complex distributed systems whose efficient operation is supported by an all-embracing communication system. In the Human brain that is the Nervous system, in the National Economics this is the Transportation system.

From Table VI, one can see that the principal features of the Transportation system are well reflected in both brain research projects. Only one feature "What is being transported?" is missing in the future brain studies. A proper answer to the question "What is being transported in the Nervous system between different brain parts?" should be "Information". But, for unknown reasons, that is left undefined in both future mega-projects. And the consequences of this omission are predictable.

TABLE V        EUROPEAN AND AMERICAN HUMAN BRAIN PROJECTS

| Parameter | European HB Project | American BRAIN Project |
|---|---|---|
| Duration | 10 years. | long-lasting programme |
| Funding | $ 1.35 billion. | $110 million in the 2014 fiscal year supposed to ramp up this commitment in subsequent years |
| Main Topics | - Human and mouse neural channelomics.<br>- Genotype to phenotype mapping the brain.<br>- Identifying, gathering and organizing neuroscience data.<br>- **Cognitive architectures.**<br>- Novel methods for rule-based clustering of medical data.<br>- Neural configurations for neuromorphic computing systems.<br>- **Virtual robotic environments, agents, sensory & motor systems.**<br>- Theory of multi-scale circuits. | - Generate a census of brain cell types<br>- Create structural maps of the brain<br>- Develop new, large-scale neural network recording capabilities<br>- Develop a suite of tools for neural circuit manipulation<br>- **Link neuronal activity to behavior**<br>- Integrate theory, modeling, statistics and computation with neuroscience experiments<br>- Delineate mechanisms underlying human brain imaging technologies<br>- Create mechanisms to enable collection of human data for scientific research<br>- Disseminate knowledge and training |

TABLE VI     JUXTAPOSING HUMAN BRAIN PROJECTS

| Economic system | Human brain system | |
|---|---|---|
| Transportation system | Nervous system | |
| System's Features | American BRAIN Project | European HB Project |
| Network topology (road and pathway maps) | The Brain Connectome Project | Neural channelomics<br>Neuroinformatics platform |
| Network dynamics (Traffic)<br>Transportation means, time tables, hubs, congestions | The DARPA's SyNAPSE Project | Neuromorphic computing platform |
| What is being transported (through the network)<br>Raw materials, Goods, Freights. | **(Information)** | **(Information)** |

On the other hand, the reason of this omission is also fully understandable: we don't know what Information is and how it is being transported (processed) in the brain. (That the brain is an information processing system is a widely accepted hypothesis in the scientific community). So, it will be wise to try to understand what information is.

## VI. WHAT IS INFORMATION

While a consensus definition of information does not exist, we would like to propose a definition of our own (borrowed and extended from the Kolmogorov's definition of information, first introduced in the mid-sixties of the past century):

**Information is a linguistic description of structures observable in a given data set** [16].

Two types of structures could be distinguished in a data set – primary and secondary data structures. The first are data elements aggregations whose agglomeration is guided by natural physical laws; the others are aggregations of primary data structures which appear in the observer's brain guided by the observer's customs and habits. Therefore, the first could be called Physical data structures, and the second, Meaningful or Semantic data structures. And their descriptions should be accordingly called **Physical Information** and **Semantic Information**.

This subdivision is usually overlooked in the contemporary data processing approaches leading to mistaken and erroneous data handling methods and techniques.

In [17], E. Diamant presents a list of publications on the subject and a more extended explanation of information description duality can be found. Meanwhile, it is important to explain the consequences that immediately pop up from this assertion. And which are critically important for the right definition of the notion of cognition. In the light of this just acquired knowledge, we can certainly posit that **cognitive ability is the ability to process information**. And that is what our brains are doing, and that is what we are striving to replicate in our Cognitive Robots designs.

First of all, physical information is carried by the data and therefore can be promptly extracted from it. At the same time, semantic information is a description of observer's arrangement of the physical data structures and therefore it can not be extracted from the data, because semantics is not a property of the data, it is a property of an observer that is watching and scrutinizing the data. As such, semantics is always subjective and it is always a result of mutual agreements and conventions that are established in a certain group of observers, or a future group of robots and humans that act as a team sharing a common semantic understanding (semantic information) about their environment. An important sequel of this is that the semantic information can not be learned autonomously, but it should be provided to a cognitive robot from the outside (semantics has to be taught and not learned, as it is usually requested by all workprogrammes).

Another important corollary that follows from the new understanding of information nature is that information description is always a linguistic description, that is, a string of symbols which can take a form of a mathematical formula (don't forget that mathematics is a sort of a language) or a natural language item – a word, a sentence, or a piece of text. That is a very important outcome of the new theory considering that contemporary approaches to the problem of information processing are assuming computer involvement in the processing task. However, contemporary computers are data processing machines which are not supposed to process natural language texts which are carrying semantic information.

Finally, we would like to provide some examples of widespread misunderstandings that appear in the Calls of proposals issued by DARPA and EU Commission: In the "ICT Work Programme 2009/2010, (C(2009) 5893)" [7], in its "Part 4.2 Challenge 2: Cognitive Systems, Interaction, Robotics" the problem that Robotic systems have to cope with is specifies as "extracting meaning and purpose from bursts of sensor data or strings of computer code…" This is a false and a misleading statement – sensor data does not possess semantics, and therefore, meaning and purpose can not be extracted from it.

DARPA's Document "Deep Learning" (RFI SN08-42) states that: "DARPA is interested in new algorithms for learning from unlabeled data in an unsupervised manner to extract emergent symbolic representations from sensory input…" Again, that is a false and a misleading statement – symbolic representations (semantics) could not be learned from data.

## VII. Conclusion

Cognitive Robotics R&D is a very important branch of contemporary science that is paving the road to the next technological revolution – smart robots that are invading our everyday lives. Until now, the extensive research efforts of the Cognitive Robotics field investigators have been derailed by a wrong understanding about the essence of information, in general, and semantic information, in particular. We hope that the paper will contribute to some changes in this situation.